\documentclass[letterpaper]{article} 
\usepackage{aaai25}  
\usepackage{times}  
\usepackage{helvet}  
\usepackage{courier}  
\usepackage[hyphens]{url}  
\usepackage{graphicx} 
\usepackage{natbib}  
\usepackage{caption} 
\usepackage{algorithm}
\usepackage{algorithmic}
\usepackage{textcomp}
\usepackage{booktabs}

\usepackage{newfloat}
\usepackage{listings}
\DeclareCaptionStyle{ruled}{labelfont=normalfont,labelsep=colon,strut=off} 
\floatstyle{ruled}
\newfloat{listing}{tb}{lst}{}
\floatname{listing}{Listing}
\usepackage{xcolor}

\title{Characterizing Knowledge Manipulation in a Russian Wikipedia Fork}
\author {
    Mykola Trokhymovych\textsuperscript{\rm 1},
    Oleksandr Kosovan\textsuperscript{\rm 2},
    Nathan Forrester\textsuperscript{\rm 3},
    Pablo Aragón\textsuperscript{\rm 1},\\ 
    Diego Saez-Trumper\textsuperscript{\rm 1},
    Ricardo Baeza-Yates\textsuperscript{\rm 1}
}
\affiliations {
    \textsuperscript{\rm 1}Universitat Pompeu Fabra\\
    \textsuperscript{\rm 2}Ukrainian Catholic University\\
    \textsuperscript{\rm 3}Independent Scholar\\
    \{mykola.trokhymovych, pablo.aragon, diego.saez\}@upf.edu,\\ o.kosovan@ucu.edu.ua, nathanforrester.research@proton.me,
    rbaeza@acm.org
}
\nocopyright

\begin{document}

\maketitle

\begin{abstract}
Wikipedia is powered by MediaWiki, a free and open-source software that is also the infrastructure for many other wiki-based online encyclopedias. These include the recently launched website Ruwiki, which has copied and modified the original Russian Wikipedia content to conform to Russian law. To identify practices and narratives that could be associated with different forms of knowledge manipulation, this article presents an in-depth analysis of this Russian Wikipedia fork. We propose a methodology to characterize the main changes with respect to the original version. The foundation of this study is a comprehensive comparative analysis of more than 1.9M articles from Russian Wikipedia and its fork. Using meta-information and geographical, temporal, categorical, and textual features, we explore the changes made by Ruwiki editors. Furthermore, we present a classification of the main topics of knowledge manipulation in this fork, including a numerical estimation of their scope. This research not only sheds light on significant changes within Ruwiki, but also provides a methodology that could be applied to analyze other Wikipedia forks and similar collaborative projects.
\end{abstract}

\section{Introduction}
\label{sec:introduction}
\footnotetext{To appear in ICWSM’25.}
Online information dissemination plays a key role in shaping public opinions and attitudes. As the the world’s largest encyclopedia
and the ninth most visited website globally,\footnote{\url{https://www.similarweb.com/website/wikipedia.org},~Accessed 12 September 2024.} Wikipedia holds 
an influential position within the web ecosystem~\cite{10.1145/3442381.3450136}. It is maintained through a collaborative community effort to become the ``sum of all human knowledge''~\cite{sutcliffe2016wikipedia}.
That is, anyone can freely edit Wikipedia content, although several policies and guidelines decided by the community must be followed to guarantee the quality and integrity of knowledge~\cite{mcdowell2020takes}. 
Given that its content empowers various applications, for example, integrating verified facts into curricula~\cite{10.1145/3289600.3291021}, fact-checking~\cite{10.1145/3459637.3481961}, or training of large language models~\cite{devlin-etal-2019-bert}, 
knowledge on Wikipedia has a major societal impact.

There are actors who are not comfortable with the content and policies of Wikipedia. For example, some states like China and Turkey have repeatedly blocked access to the website~\cite{turkey-wikipedia-block,china-wikipedia-ban}. In the case of China, the decision to censor Wikipedia was also followed by the launch of \textit{Baidu Baike}, an online encyclopedia with content in accordance with the requirements of the Chinese Government~\cite{woo2007bloomberg,china-wikipedia-ban}. Moreover, the open-source nature of \textit{MediaWiki}, the software that powers Wikipedia, has enabled the launch of alternative wiki-based encyclopedias. A well-known example is \textit{Conservapedia}, created by detractors of Wikipedia's core policy of neutrality and self-described as American conservative and fundamentalist Christian~\cite{johnson2007conservapedia}. Another example is \textit{Runiversalis}, the content of which must follow the requirements of the Russian legislation and its traditional values~\cite{runiversalis}. While the views on these encyclopedias don't compare to Wikipedia's, the growing cultural prominence of alternative facts has sparked a noticeable rise in both traffic and interest~\cite{wiredWelcomeWikipedia}.

\begin{figure}[b]
  \centering
  \includegraphics[width=\linewidth]{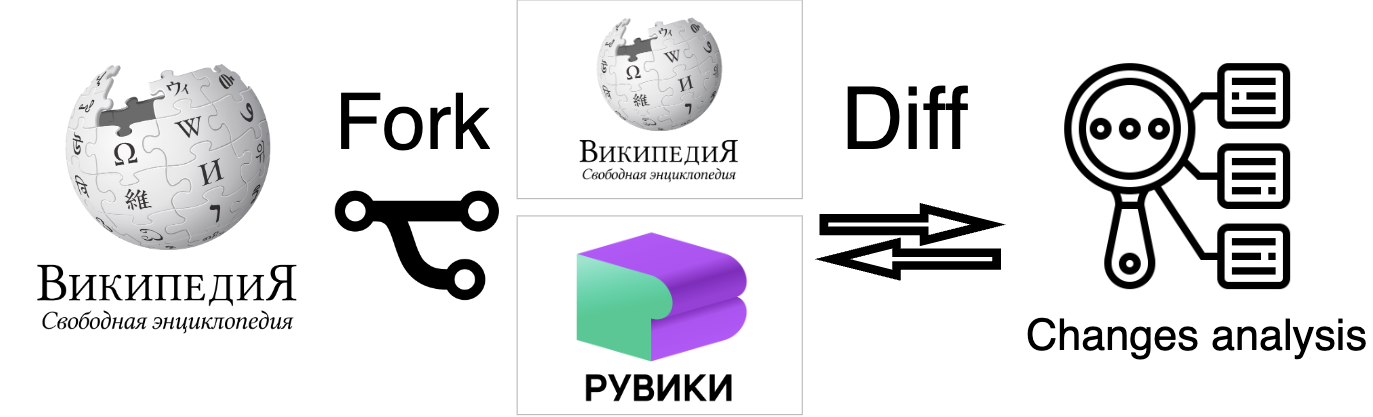}
  \caption{Summary of our research that analyzes changes in a Russian Wikipedia fork to assess knowledge manipulation.}
  \label{fig:screenshot}
\end{figure}

Russian Wikipedia appeared in May 2001 during the first wave of non-English Wikipedias. 
In June 2023, a fork
of Russian Wikipedia was launched online, hereinafter referred to as \textit{RWFork}.
The project was founded by Vladimir Medeyko, former Director of Wikimedia Russia, a Wikimedia Chapter organization.
While this project was powered by \textit{MediaWiki} software like the aforementioned alt-Wikipedias, its content was also copied from Russian Wikipedia and later edited to conform to the Russian legislation~\cite{bloomberg-2023-putin-friendly-clone}. Therefore, \textit{RWFork} is an organized effort to manipulate knowledge, originally created with neutral editorial policies, in order to comply with the editorial policies of a specific state.

\paragraph{Research Questions.} 
In this work, we examine a Russian Wikipedia fork. Our primary goal is to characterize how original content from Russian Wikipedia has been manipulated (see Figure~\ref{fig:screenshot}). To achieve this, the initial step is to identify which articles have undergone changes and their relevance. Therefore, our first research question is:
\begin{itemize}
    \item \textbf{RQ1} How relevant are Russian Wikipedia articles changed by \textit{RWFork} editors?
\end{itemize}
Next, we aim to examine changes made within \textit{RWFork} in detail to characterize the editorial process and the nature of content modifications. This leads to our second research question:
\begin{itemize}    
    \item \textbf{RQ2} How has article content changed in \textit{RWFork}?    
\end{itemize}
Finally, we are interested in categorizing the changes that have occurred in \textit{RWFork} to provide a clear understanding of the broader patterns of knowledge manipulation. As a result, our third research question is:
\begin{itemize}    
    \item \textbf{RQ3} What are the patterns of knowledge manipulation in \textit{RWFork}?
\end{itemize} 

\paragraph{Methodology.} To answer our research questions, we have developed a methodology to compare the content of two  MediaWiki-powered websites: Russian Wikipedia and \textit{RWFork}. 
The first challenge we face is data collection and preparation. 
Relying on MediaWiki APIs, we implemented a data retrieval approach to 
extract data from both sources.
To address \textbf{RQ1}, we use bootstrapping to estimate the relevance metrics of articles along with their confidence intervals. Later we use these estimations to characterize the articles that were changed by \textit{RWFork} editors. As for \textbf{RQ2}, we use exploratory data analysis on a wide range of temporal, geographical, categorical, and topical features.
Our methodology also includes named entity recognition to identify the most frequent entities in deletions and additions.
To answer \textbf{RQ3}, we combine advanced natural language processing tools, clustering algorithms, and qualitative analysis to build a classification of the main topics affected by knowledge manipulation.

\paragraph{Main Findings.} By applying our methodology to a dataset with more than 1.9M articles of the Russian Wikipedia and its fork, we discover that:
\begin{itemize}
    \item \textbf{RQ1}: \textit{RWFork} editors have modified articles that have a considerably higher number of page views compared to others, hence influential ones. We find that 1.75\% of articles (\textit{changed} pages) generate 14.2\% of the total page views. Also, according to our analysis, modified pages refer to 
    controversial
    topics, and have almost twice the higher revert rate than duplicated ones. 
    
    \item \textbf{RQ2}: Unlike Wikipedia editors, \textit{RWFork} editors have a shorter activity period that aligns with standard office hours and reduced weekend activity. Most frequently altered pages pertain to Ukraine, Russia, and Belarus locations. The most frequently added and deleted categories are related to the 2022 Russian invasion of Ukraine, particularly occupied territories or sanctions. Also, we discover frequent additions of Russian Government resources, contrasting with the frequent deletion of EU and Ukrainian Government websites.
    \item \textbf{RQ3}: We 
    find
    that most of the content changes ($\sim$90\%) can be classified into 8 main categories. In particular, we discover that the majority of the content changes are knowledge manipulations related to territory reassignment, international sanctions, and terminology variations related to the 2022 Russian invasion of Ukraine ($\sim$44\%).
\end{itemize}

To our knowledge, our work provides the first study of how original Wikipedia content has been forked and manipulated to meet the requirements of a national regulation. In addition to the geopolitical relevance of this case study, our methodology can also support future work examining the differences between wiki-based encyclopedias.
To encourage further research, we release the dataset of this study under an open license using the Zenodo platform,\footnote{\url{https://zenodo.org/records/15073728}} along with the code used for the presented case study.\footnote{\url{https://github.com/trokhymovych/RWFork}}

\paragraph{Relevance.}

National identity and public opinion can be influenced by the information citizens are finding 
online about their history. In a previous study, Wikipedia 
was ranked the 6th most important information about history, passing museum visits, college courses, and social media~\cite{aha2021snapshot}. Therefore, attempts to manipulate Wikipedia content, even if they happen in other platforms, could have a significant societal impact.

Wikipedia is also a key resource for web search engines~\cite{vincent2019measuring,vincent2021deeper}. 
Moreover, previous research has shown that Wikipedia is not only an important source of information but also has a role as a high-traffic gateway to the broader Web ecosystem~\cite{10.1145/3442381.3450136}. 
As a consequence, limitations of access to Wikipedia or replacing it with alternative versions could favor the displacement of web traffic to specific resources with manipulated information.

Last but not least, Wikipedia content is frequently used for training Large Language Models (LLMs)~\cite{devlin-etal-2019-bert}. Manipulated versions of Wikipedia used as training data for LLMs can encourage AI-powered systems that promote ideas with specific biases~\cite{10.1145/3442188.3445916}. 
Therefore, it is crucial to characterize what biases are included in \textit{RWFork} as the plan to use its data to train LLMs has already been announced by the project founder~\cite{lenta-ruviki-2023}.

\section{Related Work}
\label{sec:re_wo}

To contextualize our characterization of knowledge manipulation in \textit{RWFork}, we categorize prior research on Wikipedia into three main areas: knowledge gaps, knowledge integrity, and the specific case of the 2022 Russian invasion of Ukraine.

\subsection{Knowledge gaps}

Wikipedia aims to represent ``the sum of all human knowledge'' while retaining the requirement of the neutral point of view~\cite{10.1145/2491055.2491064}. This ambitious goal has been the subject of extensive research aimed at identifying biases in the form of knowledge gaps, {\em i.e.}, ``disparities in content coverage or participation of a specific group of readers or contributors''~\cite{redi2020taxonomy}. 

Gender gap is arguably the most extensively studied knowledge gap on Wikipedia~\cite{reagle2011gender,eom2015interactions,hinnosaar2019gender,wagner2016women,zagovora2017weitergeleitet}. However, several studies have also explored cultural and geographic disparities in topic coverage. Early research provided empirical evidence of self-focus within multiple language editions of Wikipedia~\cite{hecht2009measuring}, a phenomenon that likely contributes to the significant disparities in the geographical representation of knowledge~\cite{graham2011geographies, graham2014uneven, beytia2020positioning}. Self-focus might also play an important role in the cultural local biases observed in content across languages~\cite{hecht2010localness,callahan2011cultural,miquel2021role} and the biased narratives found on controversial historical events and cultural heritage~\cite{rogers2012neutral,pentzold2017digging}. 

Despite knowledge gaps, research has revealed that Wikipedia's content is no more biased than that in expert-written encyclopedia articles~\cite{10.25300/MISQ/2018/14084}. 
Since the plan for the Russian Wikipedia fork is to be initially edited by experts~\cite{bloomberg-2023-putin-friendly-clone}, it becomes particularly compelling to identify biases that could have arisen.

\subsection{Knowledge integrity}

As anyone can edit Wikipedia, editors dedicate substantial effort to monitor articles, improving content verifiability, and strengthening its resilience against misinformation~\cite{saeztrumper2019online}. Empirical research on knowledge integrity in Wikipedia has highlighted several threats currently being addressed.

Many instances of disinformation on Wikipedia, such as hoax articles, have been found to be identified and addressed quickly, which minimizes their impact~\cite{kumar2016disinformation}. A more pressing challenge to knowledge integrity is vandalism, a form of abuse that has drawn significant attention from research. Numerous studies have analyzed its characteristics~\cite{shachaf2010beyond,geiger2010work,potthast2010crowdsourcing} and proposed detection systems for this problem~\cite{potthast2008automatic,adler2011wikipedia,10.1145/3580305.3599823}. Detection efforts have also focused on the phenomenon of sock puppets~\cite{kumar2017army,sakib2022automated}, including cases to evade account bans~\cite{niverthi2022characterizing}.

The efforts of Wikipedia editors to preserve knowledge integrity have contributed to transforming the project from a questionable source of information in its early years into an increasingly reliable one over time~\cite{steinsson2024rule}. Recent research has highlighted that the community governance infrastructures of Wikipedia are crucial in addressing systematic disinformation campaigns and other influence operations~\cite{kharazian2023governance}. For that reason, examining the changes that have occurred in \textit{RWFork} can offer important insights into how Wikipedia knowledge could be manipulated without its community governance.

\subsection{Case: the 2022 Russian invasion of Ukraine}

The great importance of the Russian invasion of Ukraine in 2022 has led to a growing body of literature on the documentation of this specific event in Wikipedia.
A first study of the English Wikipedia article on this conflict highlighted the role of vandal fighters in facilitating coordinated editing efforts during a fast-changing and contentious event~\cite{roberts2022wikipedia}.
A later analysis of the effects of the conflict on multiple articles and languages showed a significant decline in activity around the time of the invasion on both Russian and Ukrainian language editions, followed by a recovery~\cite{dammakeffects}. Interestingly, there was a sharp increase in the rate of reverts right after the invasion. More recently, an interview study with expert editors from English Wikipedia showed no evidence of a state-sponsored information operation, although participants reported disruptive editing in war-related articles from accounts aligned with either Russian or Ukrainian positions~\cite{kurek2024wikipedia}.

All these studies highlight the critical role of Wikipedia's editorial norms in preventing state-sponsored information operations during the Russian invasion of Ukraine in 2022. As a consequence, examining a fork of Wikipedia created to comply with the Russian legislation offers a valuable opportunity to envision how its content could have been manipulated in the absence of Wikipedia editorial standards.

\begin{figure*}[t]
  \centering
  \includegraphics[width=\linewidth]{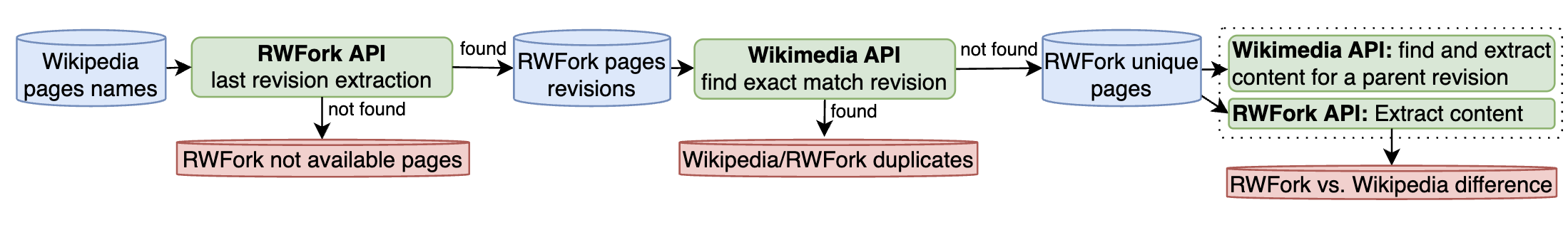}
  \caption{Process for crawling revision differences between \textit{RWFork} and Russian Wikipedia.}
  \label{fig:ruwiki_crawling}
\end{figure*}

\section{Data Collection and Preparation}
\label{sec:collection}

A major challenge of this work is data collection, as it
requires parsing 
data from two 
different \hbox{\textit{MediaWiki}}-powered websites
on a large scale and further post-processing. In this section, we present our process to collect data.

\subsection{Article Selection}
The first step of the data collection process was to define the articles of our dataset. \textit{RWFork} was initially created as a copy of Russian Wikipedia, meaning that most of the content, including page titles, is the same. 
As a consequence, we used the page title as a key to match articles from \textit{RWFork} and  Russian Wikipedia. Page titles were extracted from the existing Russian Wikipedia articles of the June 2023 Wikimedia dump.\footnote{\url{https://dumps.wikimedia.org/}}
In total, we compiled a list of about 1.9M distinct page titles for further processing. 

We should note that there are no similar resources from \textit{RWFork}. Therefore, our pipeline did not include newly created articles, which is one of the limitations of our research.

\subsection{Web Crawling}\label{sec:ruwiki_crawling}
For Russian Wikipedia, we extracted the content of pages of our dataset, formatted as wikitext,\footnote{\url{https://en.wikipedia.org/wiki/Help:Wikitext}} using the \textit{Wikimedia} API.\footnote{\url{https://ru.wikipedia.org/w/api.php}}
For \textit{RWFork}, although the project is also powered by \textit{MediaWiki} 
and provides an API, several limitations led us to collect data through a multi-step process. The full crawling pipeline is presented in Figure~\ref{fig:ruwiki_crawling}.

Our main interest 
is
to understand the contribution of \textit{RWFork} 
editors,
 i.e., 
information about what was added or deleted compared to the original 
Russian
Wikipedia version. As for that, we 
examined the
page history that includes 
edits 
(also known as 
revisions).
First, we examined if \textit{RWFork} articles were an exact copy of Russian Wikipedia at some point in time. For that purpose, we parsed the last \textit{RWFork} revision of each 
article 
to match it with the corresponding Russian Wikipedia edit history using revision and parent revision IDs as a compound key. In this step, a subset of articles were not available in \textit{RWFork}. 
If the last \textit{RWFork} revision is not included in Russian Wikipedia history, we consider that this page includes changes created by \textit{RWFork} editors. Then, using \textit{RWFork} page history, we extracted its Russian Wikipedia parent revision (the last revision of the \textit{RWFork} page that has a match in Russian Wikipedia history). Finally, we extracted the content for the \textit{RWFork} page's last version and its Russian Wikipedia parent revision
to identify the pieces of information that were modified. 

Our dataset includes records from May to September 2023.
We processed 1,925,452 pages, where 17,745 (0.92\%) were unavailable (either deleted or with limited access), and 33,664 (1.75\%) pages included \textit{RWFork}-based edits.

\subsection{Data Processing}
After collecting pairs of article versions from two data sources, the next step was to extract 
their differences. We use an open \textit{mwedittypes}\footnote{\url{https://github.com/geohci/edit-types}} library for text processing.
We considered the Russian Wikipedia article as a base version studying the modification needed to achieve the \textit{RWFork} version.

First, we extracted the sentences or phrases that were \textit{inserted} or \textit{deleted}. Those lists may contain similar items. Therefore, we define the additional category \textit{changed} by pairwise matching
sentences from the \textit{inserted} and \textit{deleted} lists using the Levenshtein Distance~\cite{levenshtein1966binary}. 
Following the methodology used previously for Wikipedia revisions comparison in~\cite{10.1145/3580305.3599823}, we adopted a similarity threshold of 0.6. When the similarity of two sentences exceeded this threshold, we moved the pair to the \textit{changed} list and removed them from the \textit{inserted} and \textit{deleted} lists. This process resulted in lists of \textit{inserted}, \textit{deleted}, and \textit{changed} sentences for each article.

Moreover, we extracted the lists of changed media names, page categories, and references. Examples of parsed content changes are included in Appendix~\ref{add:data_sample}.

\paragraph{Additional Data Sources.}
Besides the
differences comparing \textit{RWFork} and Russian
Wikipedia
article revisions, we
retrieved additional article characteristics 
necessary to address \textbf{RQ1} and \textbf{RQ2}. 
In particular, we extracted the list of countries and topics related to the article. 
We assumed that those are the same for the \textit{RWFork} and Wikipedia versions. As for the countries extraction, we used 
a tool that provides countries 
predictions for Wikipedia articles based on their associated Wikidata items and links to other Wikipedia articles.\footnote{\url{https://wiki-topic.toolforge.org/countries}} As for the topics, we 
relied on a topic prediction tool for Wikipedia articles based on their links to other articles~\cite{10.1145/3442442.3452347}.\footnote{\url{https://wiki-topic.toolforge.org/topic}}
Furthermore, we extracted the monthly number of views per page from the Wikimedia API.


\section{RQ1: Relevance of Changed Articles} 
\label{sec:rq1}
Our dataset contains information about more than 1.9M article pairs. Of those, 97.33\% of them
are just \textit{duplicated} (not changed in \textit{RWFork}) and 0.92\% are \textit{missing}. Among the 33,664 (1.75\%) of articles \textit{changed} by \textit{RWFork} editors, 0.96\% contain changes within the text and another 0.79\% only have changes in elements that do not affect the text (formatting, references, tags, media, etc.). 

Our study begins with an examination of metrics related to page relevance on Wikipedia. 
We use the 2022 and 2023 Russian Wikipedia history dump\footnote{Data includes records by October 2023} and page view statistics.
Specifically, we analyze the average number of page views per month, the number of edits, the rate of IP edits, and the revert rate (the proportion of edits identified as damaging and subsequently reverted). 
For each metric and group (\textit{changed}, \textit{duplicated}, \textit{missing}), we perform 10K bootstrap resamples of 1K page statistics each, sampled with replacement. This balances data variability, computational efficiency, and estimate reliability, enabling calculation of the mean and 95\% confidence interval using quantiles~\cite{efron1994introduction}.

\begin{figure}[t]
  \centering
\includegraphics[width=\linewidth]{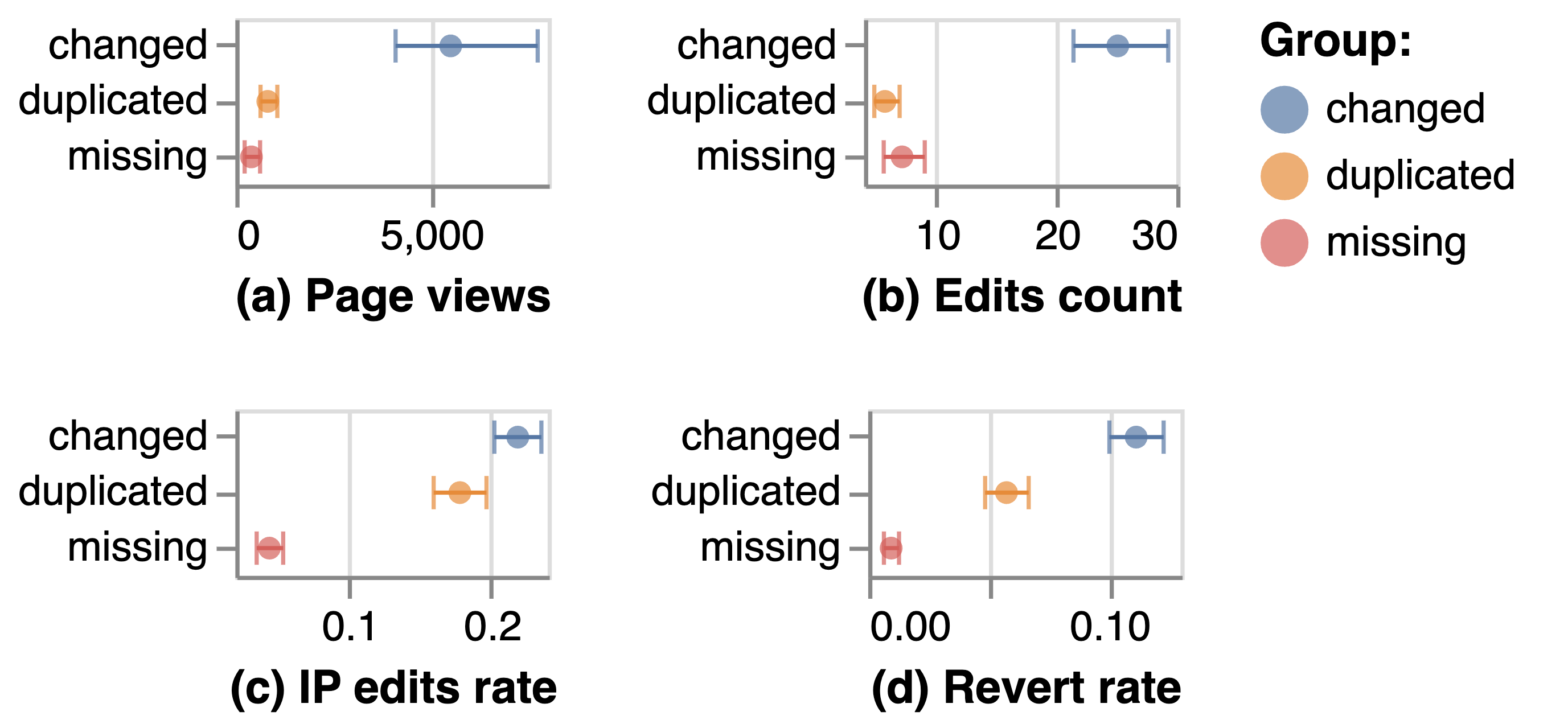}
  \caption{Comparison of Russian Wikipedia pages statistics for the groups of \textit{changed}, \textit{duplicated}, and \textit{missing} pages. Statistics used: (a)~Monthly page views; (b)~Edits count; (c)~IP edits rate; and (d)~Revert rate. Plots include mean values with 95\% confidence intervals for corresponding statistics.}
  \label{fig:relevans_stats}
\end{figure}

Results are presented in Figure~\ref{fig:relevans_stats}.
We observe that \textit{changed} pages have significantly more page views than \textit{duplicated} ones. Although only about 1.75\% of pages were changed, these pages generate approximately 14.2\% of Russian Wikipedia's page views (around 9.6\% from pages with text changes and 4.6\% from those with other changes), indicating their popularity.
Also, articles that were changed in \textit{RWFork} have significantly more edits and IP edits in Russian Wikipedia than \textit{duplicated} ones, 
an indicator of a higher attention level from registered and unregistered editors. 
Finally, the revert rate of \textit{changed} articles is almost twice higher than for \textit{duplicated} ones, a signal of 
higher risk of disputes and vandalism for those articles~\cite{10.1145/3580305.3599823}.

\section{RQ2: Changes of Article Content} 
\label{sec:EDA}

In this section, we explore article changes by comparing the general characteristics of the Russian Wikipedia and \textit{RWFork}, including temporal, geographical, categorical, source, media, and text-based features.

\paragraph{Editing Time.}

\begin{figure}[t]
  \centering
  \includegraphics[width=0.99\linewidth]{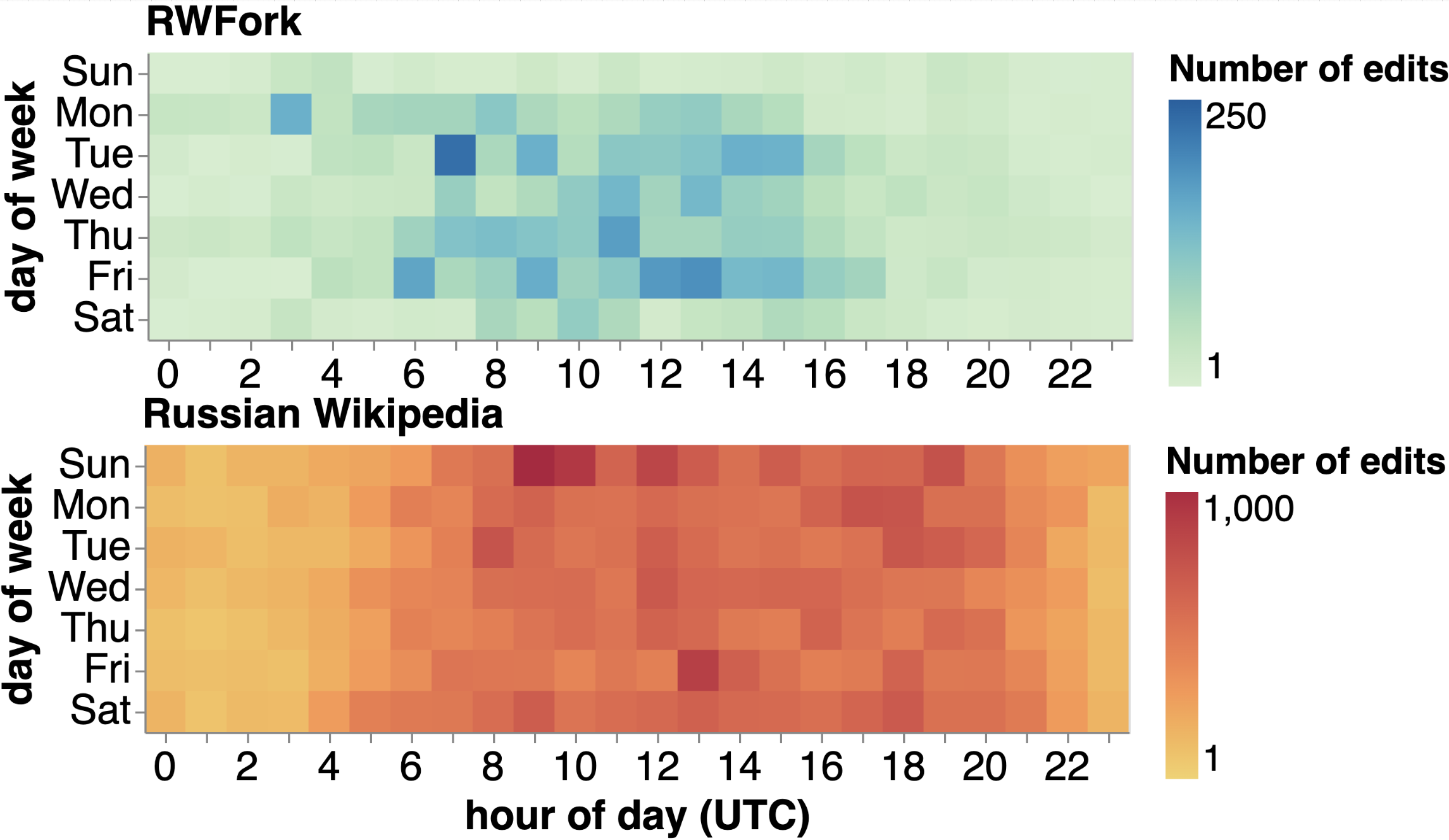}
  \caption{Average number of edits per day of week and hour of day in \textit{RWFork} (\textit{top/blue}) and Russian Wikipedia (\textit{bottom/red}). The color intensity indicates the volume of edits, with darker shades representing higher activity.}
  \label{fig:temporal_edit}
\end{figure}

We compare the temporal regularities of editing in \textit{RWFork} and Russian Wikipedia from August 2023. To reduce noise, we ignore all revisions created by bots, using a hard filter based on the username. Our findings are presented in Figure~\ref{fig:temporal_edit}. Previous research has demonstrated that Wikipedia editorial activity has circadian patterns~\cite{10.1371/journal.pone.0030091}. Russian Wikipedia follows a strict daily pattern, with a short inactivity period at night. In contrast, \textit{RWFork} editor's activity period is shorter and coincides with standard office hours, having very reduced activity during the weekend. In particular, 53.24\% of edits on \textit{RWFork} are made on weekdays from 8 to 17 UTC time, while only 40.06\% of edits on Russian Wikipedia are made in that time interval.

\begin{figure}[t]
  \centering
  \includegraphics[width=0.90\linewidth]{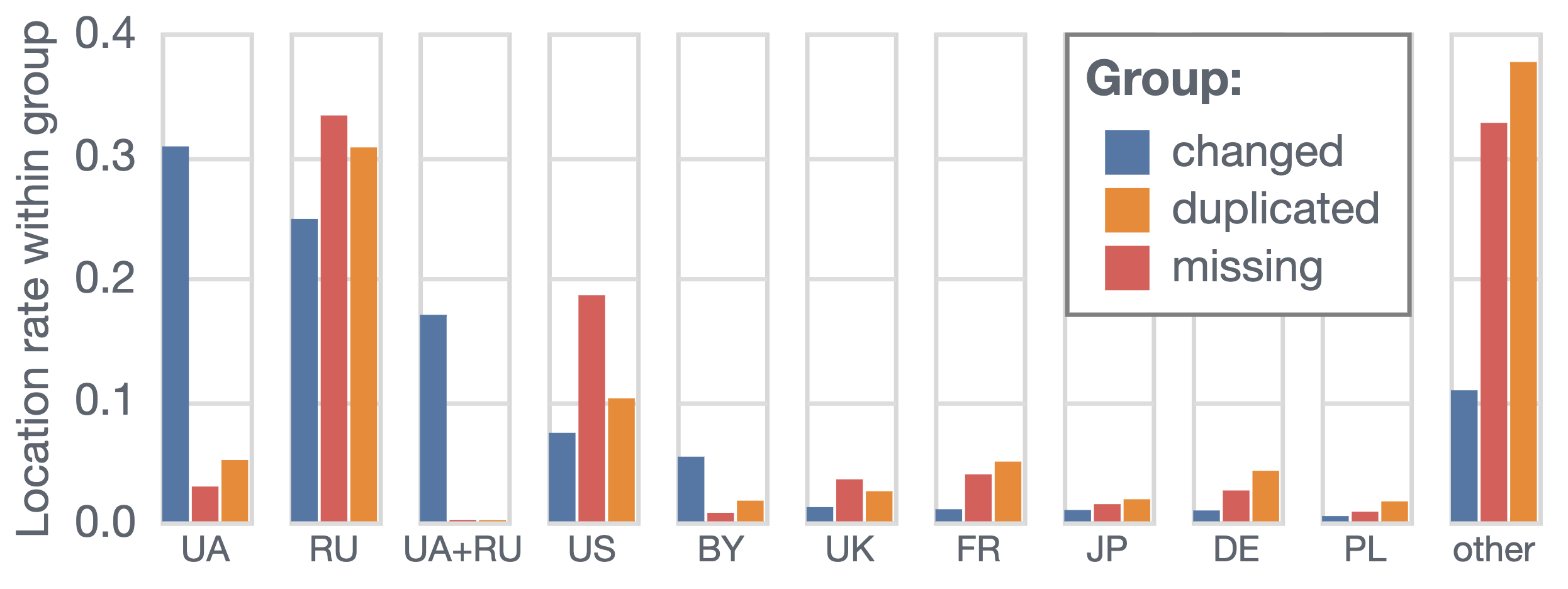}
  \caption{Rates within groups of \textit{changed}, \textit{duplicated}, and \textit{missing} pages in \textit{RWFork} for the top 10 most frequent countries in \textit{changed} group.}
  \label{fig:geo_edit}
\end{figure} 

\paragraph{Geography.}
We then analyze the geography of articles that were \textit{changed} in \textit{RWFork} compared to Russian Wikipedia ones. Although articles typically relate to one country, some relate to multiple countries (e.g., articles about disputed territories or people with links to more than one country) or no countries (e.g., pages about common knowledge topics). 
We 
also
analyze locations of pages that are either full \textit{duplicates} or \textit{missing} in \textit{RWFork}. It should be noted that we are limited to the pages that have at least one location linked, which is 53.7\% from the complete set. 
We compute the rate of pages 
related 
to the specific list of locations within  \textit{changed}, \textit{duplicated}, and \textit{missing} groups. 

The 10 most frequent locations from the \textit{changed} group the frequency rates from other groups 
are presented in Figure~\ref{fig:geo_edit}.
We find that pages from Ukraine (UA) and Ukraine+Russia (UA+RU) have a larger frequency in the \textit{changed} group than  in other groups. We also observe a similar tendency for pages related to Belarus (BY). It should be noted that 77.5\% pages from the Ukraine+Russia location have changed when the same rate for general distribution is only 1.75\%. Additionally, 
there is a remarkably high  
rate of US location in the \textit{missing} group.

\paragraph{Categories.}
We analyze categories that were added and deleted while editing \textit{RWFork} pages. In particular, for each changed page, we calculated the difference between sets of Wikipedia and \textit{RWFork} categories. 
We find 1,056 unique categories added and 991 deleted. The most frequently added and deleted categories, along with the rate of changed pages, including presented edits, are shown in Table~\ref{tab:categories}. We observe that the predominant added categories are related to the occupied territories of Ukraine. Conversely, the most frequently excluded categories relate to edit notices, individuals, and companies subject to sanctions over Russia's invasion of Ukraine. 
Despite the informative insights of these categories receiving editing focus in \textit{RWFork}, their partially disorganized structure presents challenges for automated knowledge representation. In Section~\ref{sec:clustering}, we will use these category changes to provide a structured classification of knowledge manipulation.

\begin{table}[bt]
  {\tabcolsep=1mm
  \begin{tabular}{p{6.2cm}|r|r}
    \hline
    \textbf{Added category} & 
    \textbf{Count} & \textbf{(\%)} \\
    \hline
    Villages of the Donetsk People's Republic & 621 & 1.84\\\hline
    Urban-type settlements of the Donetsk People's Republic &  213 & 0.63 \\\hline
    Urban settlements of the Donetsk People's Republic &  138 & 0.41 \\\hline
    Russian military formations that participated in military operations in Ukraine (since 2022) &  115 & 0.34 \\\hline
    Urban-type settlements of the Lugansk People's Republic & 109 & 0.32 \\
    \hline
    \hline
    \textbf{Deleted category} & 
    \textbf{Count} & \textbf{(\%)} \\
    \hline
    Articles with edit notice about caution when editing & 5,042 & 14.97 \\\hline
    Persons subject to sanctions in connection with the conflict in Ukraine &  1,412 & 4.19 \\\hline
    Temporarily occupied territories of Ukraine &  920 & 2.73 \\\hline
    Companies sanctioned over Russia's invasion of Ukraine &  452 & 1.34 \\\hline
    Urban-type settlements of Donetsk region & 129 & 0.38 \\
    \hline
  \end{tabular}
  \caption{Count and percentage of the top 5 added and deleted categories by \textit{RWFork} editors (original Russian category titles are translated to English).}
  \label{tab:categories}
  }
\end{table}

\begin{figure}[t]
  \centering
  \includegraphics[width=\linewidth]{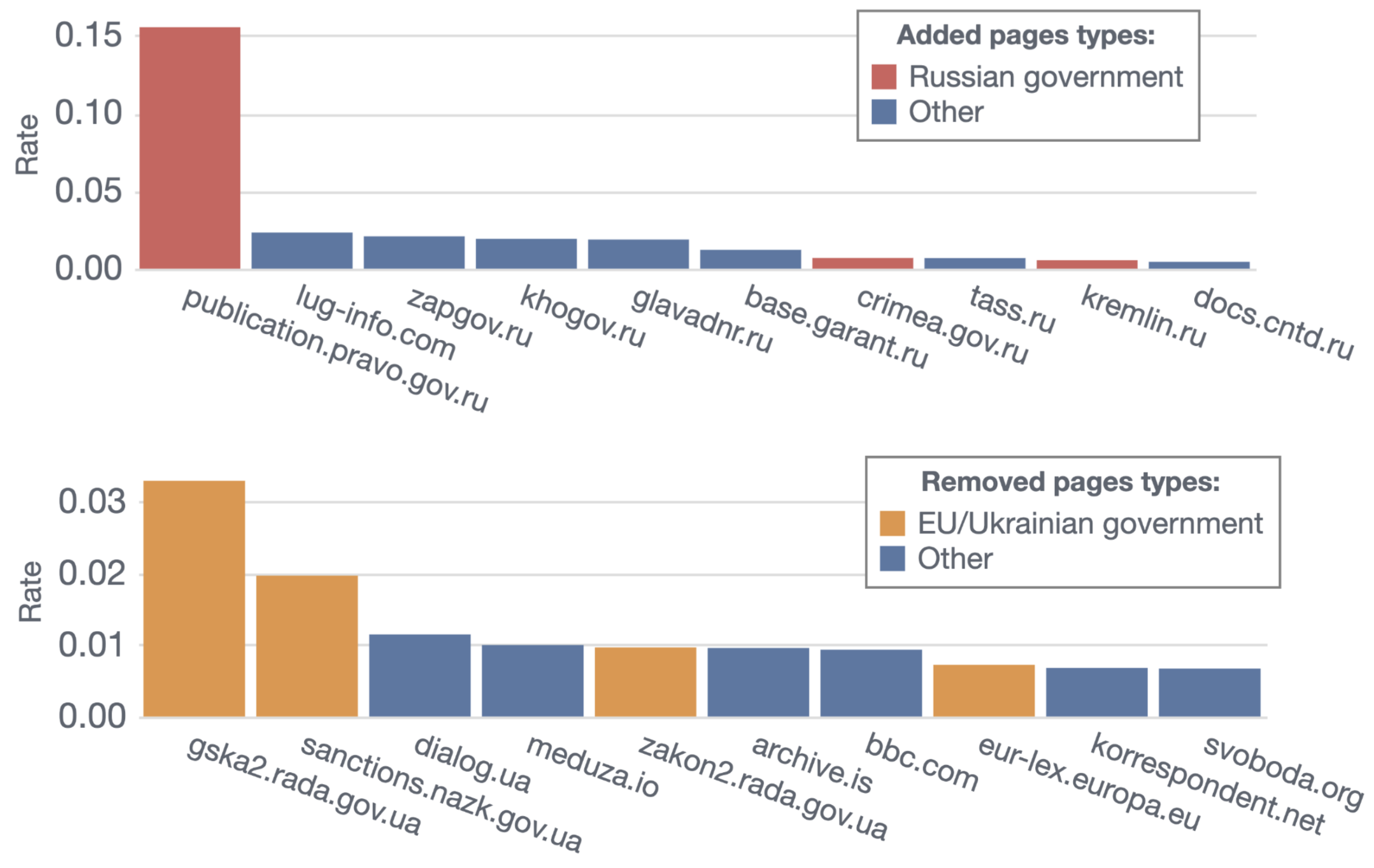}
    \caption{Top 10 most frequently added (top) and deleted (bottom) reference sources by \textit{RWFork} editors.}
  \label{fig:references}
\end{figure}

\paragraph{Sources.}
A core
content 
policy of Wikipedia
is verifiability,\footnote{\url{https://en.wikipedia.org/wiki/Wikipedia:Verifiability}} which obliges contributors to support their edits with appropriate references. This practice assures that  
readers
can verify the sources from which the information has been derived~\cite{10.1145/3366423.3380300,wiki-verifiability}.   
We analyzed reference changes to identify the sources that received the most attention from \textit{RWFork} editors. Figure~\ref{fig:references} shows 
the most added and deleted sources. On the one hand, the most frequently added sources are related to the Russian Government or administrations of occupied territories. On the other hand, the most frequently deleted sources are related to Ukrainian Government websites (e.g., the second most frequently deleted resource includes sanctions-specific information).  

\paragraph{Named Entities.}
We use 
the open-source software library \textit{SpaCy} with 
the
Russian text corpus \textit{ru\_core\_news\_sm} for named entity recognition~\cite{spacy2020}
to build two lists of named entities -- deleted and added by \textit{RWFork} editors. 
Each named entity is counted only once per page observed. 
For each named entity, we define category labels and lemma for deduplication.

Table~\ref{tab:ner-summary-stat} shows the most frequently encountered named entities that were either deleted or added. Most of these entities refer to locations, particularly those in Ukraine and Russia. We observe a tendency for \textit{RWFork} to change names to Kremlin-compliant terms for occupied territories, such as replacing ``Donetsk Oblast" to ``Donetsk People’s Republic" or its abbreviation ``DPR''. Overall, \textit{RWFork}'s modifications of named entities represent territory reassignment.

\begin{table}
  \tabcolsep=1.7mm
  \begin{tabular}
  {p{4.7cm}|p{0.75cm}|r|r}
  \hline 
    \textbf{Deleted named entities} & \textbf{Label} & \textbf{Count} & \textbf{(\%)} \\
    \hline 
    Russia & LOC & 4,169 & 22.52 \\\hline

    Ukraine & LOC & 3,956 & 21.37  \\\hline

    Verkhovna Rada & ORG & 2,358 & 12.74  \\\hline

    Donetsk Oblast & LOC & 832 & 4.49  \\\hline

    RF & LOC & 831 & 4.49  \\\hline

    Autonomous Republic of Crimea & LOC & 812 & 4.39  \\\hline

    Crimea & LOC & 721 & 3.89  \\\hline

    Luhansk Oblast & LOC & 620 & 3.35 \\\hline



    \hline
    \hline
    \textbf{Added named entities} & \textbf{Label} & \textbf{Count} & \textbf{(\%)} \\
    \hline
    Russian Federation & LOC & 4,127 & 22.29 \\\hline
    Donetsk People's Republic & LOC & 1,598 & 8.63 \\\hline
    DPR & LOC & 1,111 & 6.00 \\\hline
    Luhansk People's Republic & LOC & 1,100 & 5.94 \\\hline
    Ukraine & LOC & 1,068 & 5.77 \\\hline
    LPR & LOC & 879 & 4.75 \\\hline
    Russia & LOC & 625 & 3.38 \\\hline
    Crimea Republic & LOC & 334 & 1.8 \\
    \hline
  \end{tabular}
    \caption{Count and percentage of the top 8 added and deleted named entities by \textit{RWFork} editors (original Russian names are translated to English).}
    \label{tab:ner-summary-stat}
\end{table}

\section{RQ3: Taxonomy of Changes} 
\label{sec:clustering}

In this section, we build a taxonomy of patterns of knowledge manipulation.  
Our approach involves a comparative analysis of original articles from Russian Wikipedia and their modified versions. 
The pipeline consists of three main steps: \textit{(1)} preliminary clustering; \textit{(2)} clustering quality evaluation; and \textit{(3)} cluster correction.

\subsection{Preliminary Clustering}

To identify groups of similar edits, we first represent each revision as a single string containing the most common change types. This representation includes features such as deleted and added sentences, pairs of changed sentences, the article title, and modifications in metadata — such as categories, tags, and templates, which are essential for defining the nature of the edit. It should be mentioned that those features cover about 91\% of edits (30,599 articles), while others are omitted in further analysis. 

The strings that capture these specific edits can vary significantly in length and content. Furthermore, they often contain significant noise, including non-factual changes, irrelevant context, and parsing errors. To reduce this noise, we employ a generative model to create a fine-grained summary of each edit. Specifically, we use the OpenAI model \textit{GPT-4o-mini-2024-07-18}, with default parameters and a temperature setting of zero. 
The model was selected for its efficiency, cost-effectiveness, strong performance in Natural Language Understanding (NLU) tasks, and independence from in-house infrastructure~\cite{openai2024gpt4o}.
We provide an explicit prompt instructing the model to produce a short summary that highlights specific factual changes, constrained to a maximum length of 40 words.

Once we have the summaries of the changes, we proceed to calculate text embeddings. We use the OpenAI model \textit{text-embedding-3-small} to generate vector representations for each summary. This model produces embeddings of size 1536, with a default normalization to a magnitude of 1. These embeddings are subsequently employed for clustering using the $k$-means algorithm, as suggested in previous research~\cite{PETUKHOVA2025100}.
To determine the optimal number of clusters, we apply silhouette analysis, which helps identify the clustering configuration that maximizes the silhouette score. In our analysis, the optimal number of clusters is found to be 8, as indicated by the peak silhouette score. 

The final step in defining the taxonomy of changes involves characterizing the identified clusters. For this objective, we again use the OpenAI model \textit{GPT-4o-mini-2024-07-18}. For each cluster, we prompt the model to generate a name and a brief description, supplementing the query with a sample of 20 cluster-specific edit summaries. 
Generated description prompt to outline the specific changes made within the edits, including examples of the editing tactics employed. 

\subsection{Classification Quality Estimation}

The previous experiment produced a taxonomy of specific types of changes to Russian Wikipedia articles. This classification was created by unsupervised modeling and each cluster was characterized based on a limited number of samples. Therefore, it is essential to evaluate how accurately each sample aligns with its respective cluster name and description.
 
To address this, we employ the \textit{GPT-4o-mini-2024-07-18} model for prompt-based zero-shot binary classification, inspired by previous research~\cite{wan2024tntllmtextminingscale}, which reported a 'high agreement' with human raters for a similar task. We use the cluster name and description along with the sample summary, prompting the model to predict either yes (the sample aligns with the cluster name and description) or no. Details of the prompts used can be found in the Appendix~\ref{add:prompts}.
Our analysis reveals that 78.1\% of the samples were initially correctly classified. Table~\ref{tab:zero_shot_fit} presents the edit-to-cluster fit rate (ECFR) for each cluster along with the confidence interval (CI) for these estimates, calculated using bootstrapping, using the same approach as in Section~\ref{sec:rq1}.

Our experiment demonstrates that the majority of clusters show strong alignment between the elements and corresponding generated cluster names and descriptions. Specifically, six out of the eight clusters demonstrate an edit-to-cluster fit exceeding 87\%. However, one cluster shows a moderate fit at 59.4\% (Sanctions Edit Adjustments), while another exhibits a poor fit (LGBT Rights and History). 
Given that approximately 22\% of all samples were initially classified as not fitting their assigned group names and descriptions, we recognized the need to implement a cluster correction process, which we address in the following section.

\begin{table}[tb]
\begin{center} 
  {\tabcolsep=1.5mm
  \begin{tabular}{p{4.4cm}|c|c}
    \hline
    \textbf{Category} & \textbf{ECFR(\%)} &  \textbf{±CI(\%)}\\
    \hline
    Russian Legislation Medicines & 99.9 & 0.1 \\
    Editing Caution Removal  & 99.9 & 0.1 \\
    Cultural Metadata Updates &  99.6 & 0.2 \\
    Terminology Changes Ukraine &  97.3 & 0.6 \\
    Metadata Updates  & 92.9 & 0.7 \\
    Territorial Claims Dispute &  87.6 & 0.8 \\
    Sanctions Edit Adjustments &  59.4 & 1.6 \\
    LGBT Rights and History & ~~1.6 & 0.4 \\
    \hline
  \end{tabular}
  \caption{Edit-to-cluster fit (ECFR) before cluster correction.}
  \label{tab:zero_shot_fit}
  }
\end{center}
\end{table}

\subsection{Cluster Correction}

In this section, we explain the process of redefining clusters for initially misclassified samples using zero-shot multi-class classification applied to the previously defined taxonomy.
Specifically, we prompted the \textit{GPT-4o-mini} model 
to match the edit summary to a relevant class, providing cluster names and descriptions in the prompt. Also, we added a new category titled ``Other Changes" to allow the model to return this option when an edit does not align with any of the provided classes. 

As previously indicated, the most problematic cluster was ``LGBT Rights and History." We hypothesized that the issue arose from the cluster's name and description lacking sufficient generality. We believed that this cluster primarily contained unique changes across various topics, making it difficult to group them into a distinct category. Our correction procedure confirmed this assumption: approximately 65\% of the misclassified samples from this cluster were reassigned to the ``Other Changes" category, with the remaining distributed among other existing classes. We repeated the evaluation procedure described in the previous section and concluded that the proposed cluster correction increased the ECFR to 92\%.

\begin{table*}[t]
  {\tabcolsep=1mm
\begin{tabular}{p{2.6cm}|p{13.4cm}|c}
\hline
\textbf{Name} & \textbf{Description} &  \textbf{Size} \\
\hline
   Territorial Claims Dispute & Edits reflect changes in territorial designations and governance, emphasizing claims by the Donetsk and Luhansk People's Republics while removing Ukrainian references and administrative details. &     24.19\% \\\hline
             Metadata Updates &                                     Various edits focused on updating metadata templates, removing outdated references, and refining geographical classifications across multiple Wikipedia pages. &     18.24\% \\\hline
    Cultural Metadata Updates &                                                                         Edits focused on updating metadata with locations, cultural topics, and adding age and gender templates for various pages. &     11.77\% \\\hline
  Terminology Changes Ukraine &                           Edits focus on altering terminology related to the Russia-Ukraine conflict, shifting from specific invasion references to broader military actions and general policies. &     11.24\% \\\hline
      Editing Caution Removal &                                                             Multiple Wikipedia pages had the editing caution category removed, indicating a change in the perceived necessity for careful editing. &     10.37\% \\\hline
   Sanctions Edit Adjustments &        The edits focus on removing specific references to the Ukraine conflict in sanctions descriptions, simplifying statements, and altering context around individuals and entities sanctioned. &      8.23\% \\\hline
Russian Legislation Medicines &                                                        Templates for Russian legislation and medications were added to various pharmaceutical pages, enhancing their categorization and relevance. &      5.32\% \\\hline
      LGBT Rights and History &                              The edits focus on updating and clarifying information related to LGBT rights, historical events, and notable figures, while removing outdated or derogatory content. &      0.48\% \\\hline
                Other changes &                                                                                                                                                The edit does not fit any of the provided clusters. &     10.16\% \\\hline
\end{tabular}
  \caption{Final taxonomy of changes. Cluster names, descriptions, and sizes.}
  \label{tab:taxonomy}
  }
\end{table*}

The final taxonomy  of \textit{RWFork} changes along with their quantitative measurements are presented in Table~\ref{tab:taxonomy}. We observe that the most frequent changes refer to territorial reassignments, accounting for 24.19\% of all edits. This category represents shifts of occupied territorial entities from Ukraine to Russia. Additionally, significant groups related to the 2022 Russian invasion of Ukraine include ``Terminology Changes Ukraine" and ``Sanctions Edit Adjustments", representing 11.24\% and 8.23\% of the total edits, respectively. These clusters reflect modifications aimed at setting specific narratives, such as the removal of terms like ``invasion" and ``war", as well as adjustments to information related to sanctions across various contexts.

There are two groups that mostly consist of non-textual edits: ``Editing Caution Removal'' and ''Russian Legislation Medicines''. These categories refer to automated changes that either remove specific edit notice categories or add tags related to custom legal information. The ``Cultural Metadata Updates'' cluster consists of edits related to locations and cultural, sexual, and gender-related topics. These modifications, for example, involve specific labeling with a \textit{(+18) tag} on the pages that refer to explicit content (adult films and actors). Also, we detected a small cluster of edits that were related to the topics of ``LGBT Rights and History''.

\section{Discussion}
\label{sec:conc}

In this paper, we have presented an empirical analysis of knowledge manipulation in Wikipedia. Previous research already explored knowledge gaps in Wikipedia~\cite{redi2020taxonomy} and highlighted cultural biases across various language editions~\cite{hecht2009measuring,hecht2010localness,callahan2011cultural,rogers2012neutral,pentzold2017digging, miquel2021role}. However, our findings come from a distinct editorial process: the creation of a new platform that copied original Wikipedia content, which is then manipulated to meet the requirements of a national regulation. 
As a result, \textit{RWFork} also differs from previously studied wiki-based encyclopedias like \textit{Conservapedia}~\cite{johnson2007conservapedia}, which was created from scratch.

The proposed study can be effectively replicated in other Wikipedia forks or collaborative platforms. Examples of such forks include \textit{Runiversalis}, a wiki-based encyclopedia aligned with traditional values, and \textit{Hamichlol}, a censored wiki-based encyclopedia project for the Haredi community, among others. The methodology's adaptability lies in its ability to identify and categorize differences driven by the unique editorial policies of each wiki-based fork. 

The first step of our study focused on the relevance of Russian Wikipedia articles changed by \textit{RWFork} editors \textbf{(RQ1)}. Our analysis revealed that although the proportion is relatively small, there are articles receiving remarkable attention from readers on Russian Wikipedia. Furthermore, the articles that were altered in \textit{RWFork} receive more edits and reverts in Russian Wikipedia than those that remained unchanged. Building on previous research that has used editing and reverting activities to identify controversial topics on Wikipedia~\cite{pentzold2017digging,yasseri2012dynamics}, our observations suggest that manipulation may have taken place in popular and contentious articles. This has important implications, as controversy itself is not necessarily a negative indicator of article content. In fact, \citet{shi2019wisdom} found that Wikipedia articles edited by polarized groups of contributors typically exhibit higher quality. Therefore, if popular and controversial articles on Russian Wikipedia are forked to be edited in alignment to Russian legislation, their quality is expected to be affected.

We have then conducted a thorough analysis of how the content of articles changed in \textit{RWFork} \textbf{(RQ2)}. 
The analysis of editing time preferences shows that \textit{RWFork} are more likely to be active during standard office hours than Russian Wikipedia editors. As the initial plan for this platform is to rely on experts~\cite{bloomberg-2023-putin-friendly-clone}, one possible explanation could be that much \textit{RWFork} editorial activity is driven by paid workers. Alternatively, the geographical distribution of editors could also contribute to these differences, with Russian Wikipedia editors being more dispersed. This interpretation is consistent with the findings of~\citet{yasseri2012circadian}, who indicated that although most native Persian speakers reside in Iran, a significant portion of editing activity in the Persian Wikipedia originates from communities outside the country.
The remainder of the analysis -- encompassing article geography, categories, sources, and named entities -- reveals a clear trend: most changes are related to the 2022 Russian invasion of Ukraine. Previous research has shown that Wikipedia coverage can be influenced by the community's self-focus~\cite{hecht2009measuring} while also highlighting the success of Wikipedia communities in preventing state-sponsored information operations in articles about this conflict~\cite{roberts2022wikipedia,dammakeffects,kurek2024wikipedia}.
Therefore, there may be a link between this specific topic, as a core focus of knowledge manipulation within \textit{RWFork}, and the project founder's declared goal of ensuring compliance with Russian regulatory requirements~\cite{bloomberg-2023-putin-friendly-clone}.

Our last effort was focused on building a taxonomy of patterns of
knowledge manipulation in \textit{RWFork} \textbf{(RQ3)}.
To achieve this goal, we developed a robust clustering pipeline that incorporates intermediate steps designed to ensure the quality of the process. Although many clusters are associated with the 2022 Russian invasion of Ukraine, other topics of social importance also emerge, such as ``Russian Legislation on Medicines'' and ``LGBT Rights and History''. As previously noted, Wikipedia content is extensively used as a primary resource for LLMs~\cite{devlin-etal-2019-bert}. Studies have shown that AI systems can form significantly different associations between adjectives and political concepts based on whether they are trained on Wikipedia content or on content from web encyclopedias subject to national regulations~\cite{10.1145/3442188.3445916}. In light of the recently announced plans to integrate \textit{RWFork} into the training process for future LLMs~\cite{lenta-ruviki-2023} and the increasing societal impact of AI, our study seeks to raise awareness of the critical importance of closely examining the quality, neutrality, and potential biases of knowledge repositories.

\paragraph{Limitations and Future Work.} 
The \textit{RWFork} platform is relatively new. It is actively running and regularly introducing new content modifications. Consequently, this ongoing process may lead to the emergence of new types of changes. Also, \textit{RWFork} provides limited access to data compared to Russian Wikipedia. Therefore, our analysis did not include \textit{RWFork} newly created pages. 
It might result in missing other, undiscovered types of knowledge manipulation, but since creating new pages requires more resources, their likely limited number suggests minimal impact on our findings. Future work could address this by finding those pages through parsing \textit{RWFork} internal links between articles~\cite{10.1145/3487553.3524930}. Also, we rely on the assumption that \textit{RWFork} pages have similar geographical, categorical, and topical features as in Russian Wikipedia. Moreover, \textit{RWFork} is only one of  
several MediaWiki-powered websites. 
We therefore plan to replicate this study 
with alternative encyclopedias in future work.

Additionally, we're experimenting with a specific and limited set of models for summarization, zero-shot classification, embeddings, and clustering. We acknowledge that using different or more advanced models could improve our results, and we consider exploring this in future work.

\paragraph{Ethics.}
In this work, we have only collected and analyzed openly available data. 
Our methodology does 
not include techniques to identify users or link profiles across platforms.
The data collected does not include any private information. 
We confirm that we have read and abided by the code of conduct.

\section*{Acknowledgments}
The work of Mykola Trokhymovych is funded by MCIN/AEI /10.13039/501100011033 under the Maria de Maeztu Units of Excellence Programme (CEX2021-001195-M).

\bibliography{main}

\appendix

\section{Content Changes Examples}
\label{add:data_sample}
In Table~\ref{tab:examples}, we present a representative sample of the collected data, showing various types of content changes made to different pages.

\section{Technical Details}
\label{add:prompts}
Here we provide additional technical details to help interpret the results and improve reproducibility. Specifically, we present the prompt used for text summarization (see Figure~\ref{prm:summary}), edit-cluster alignment evaluation (see Figure~\ref{prm:eval}), and cluster reassignment (see Figure~\ref{prm:class}).

\begin{table*}[t!]
\begin{tabular}{p{4.8cm}|p{12.0cm}}
\toprule 
\textbf{Page title} & \textbf{Content changes} \\
\midrule
Zburevsky Kut
 & \textbf{Lines changed:} [\\
 &("Zburevsky Kut or Zburevsky Liman is a lake, bay in the Dnipro delta, located on the territory of the Skadovsky district \textit{(Kherson region, Ukraine)} \\
 & "Zburevsky Kut or Zburevsky Liman is a lake, a bay in the Dnipro delta, located on the territory of the Skadovsky district \textit {Kherson region of Russia}")] \\
\midrule
Medal "For the Return of Crimea"
 & \textbf{Categories removed:} ["Articles with edit notes about caution when editing"] \\
\midrule
Vladimir Samokish
 & \textbf{Lines changed:} \\
 & [("\textit{Due to Russia's invasion of Ukraine}, he is \textit{under international sanctions} of the European Union, USA, Great Britain, and other countries", \\
  & "He is \textit {on the sanctions list} of the European Union, USA, Great Britain, and other countries")] \\
  & \textbf{Categories changes:} ["Persons under the sanctions related to the conflict in Ukraine"] \\
\midrule
Trimetozine
 & \textbf{Tag added}: Legislation of the Russian Federation|Medicines \\
\midrule
Kendra (name) 
 & \textbf{Lines deleted:} ["1981) - American porn actress and erotic model", "Sunderland, Kendra (born | 1995) - American porn actress and erotic model"] \\
\midrule
Svoya igra & \textbf{Lines changed:} \\
 &[("Broadcast on NTV on \textit{Sundays} at 15:00", \\
  & "Broadcast on NTV on \textit{weekends} at 15:00")] \\
\bottomrule
\end{tabular}
\caption{Sample of content changes of various types made by \textit{RWFork} editors (original Russian texts translated to English).}
\label{tab:examples}
\end{table*}

\begin{figure*}[t]
    \centering
    \begin{lstlisting}    
    You will be provided with details regarding edit to the Wikipedia page.
    You need to deeply analyse the changes, define what was edited and provide a description of the changes.
    
    Provide a short summary and specific factual changes. 
    Pay attention to details about adding/removing/changing characteristics.
    Avoid generalizations and provide specific examples.  (max {MAX_WORDS} words)
    Return the answer in JSON format with only "desc" field and the following structure:
    {{
        "desc": "string"  # description of specific factual changes
    }}
    The edit to analyze will be provided in the <>: <{EDIT_STRING}>
    \end{lstlisting}
    \caption{Prompt template used to define the summary for content changes.}
    \label{prm:summary}
\end{figure*}

\begin{figure*}[t]
    \centering
    \begin{lstlisting}    
    You are provided with a specific edit to the Wikipedia page (defined in <>) along with possible cluster details (defined in ~~) to which the edit belongs.
    You need to analyse the edit and decide whether the edit fits the provided cluster or not.
    Provide ONLY a short answer (YES or NO).
    Edit summary: <{EDIT_SUMMARY}>
    Cluster details: ~{CLUSTER_DETAILS}~
    \end{lstlisting}
    \caption{Prompt template used to define the edit-cluster alignment.}
    \label{prm:eval}
\end{figure*}

\begin{figure*}[t!]
    \centering
    \begin{lstlisting}    
    You are provided with a specific edit to the Wikipedia page (defined in <>).
    You need to reclassify the edit to the correct cluster based on the provided cluster details.
    Cluster details:
    {ALL_CLUSTERS_DETAILS}
    8. Other changes: The edit does not fit any of the provided clusters. (always use this option if the edit does not fit any of the provided clusters)

    Provide ONLY a short answer (cluster number).
    Edit summary: <{EDIT_SUMMARY}>
    \end{lstlisting}
    \caption{Prompt template used for cluster reassignment.}
    \label{prm:class}
\end{figure*}

\end{document}